%% file: main.tex
\documentclass[conference]{IEEEtran}
\usepackage{amsmath,amsfonts}
\usepackage{algorithm}
\usepackage{array}
\usepackage{textcomp}
\usepackage{stfloats}
\usepackage{url}
\usepackage{verbatim}
\usepackage{graphicx}
\usepackage{cite}
\usepackage{xcolor}
\usepackage{adjustbox}
\usepackage[symbol]{footmisc}

\definecolor{neonblue}{rgb}{0.1, 0.7, 1}
\definecolor{neongreen}{rgb}{0.1, 1, 0.1}
\input{include/packages}

\input{include/macros}
\IEEEoverridecommandlockouts

\begin{document}

\title{ROMAN: Open-Set Object Map Alignment for Robust View-Invariant Global Localization}

\author{
\IEEEauthorblockN{Mason B. Peterson\textsuperscript{1},
Yixuan Jia\textsuperscript{1}, Yulun Tian\textsuperscript{2}, Annika Thomas\textsuperscript{1}, and
Jonathan P.\ How\textsuperscript{1}}

\thanks{This work is supported in part by the Ford Motor Company, DSTA, ONR, and ARL DCIST under Cooperative Agreement Number W911NF-17-2-0181.}
\thanks{\textsuperscript{1}Massachusetts Institute of Technology, Cambridge, MA 02139, USA. \{\texttt{masonbp, yixuany, annikat, jhow}\}\texttt{@mit.edu}.}
\thanks{\textsuperscript{2}University of California San Diego, San Diego, CA 92093, USA.
\texttt{yut034@ucsd.edu}}
}

\maketitle

\input{paper/abstract}

\input{paper/introduction}

\input{paper/related_works}

\input{paper/roman}

\input{paper/data_association}

\input{paper/mapping}

\input{paper/experiment}

\input{paper/ablations}

\input{paper/limitations}
\input{paper/conclusion}

\bibliographystyle{IEEEtran} %
\bibliography{references}

\end{document}

%% file: include/packages.tex
\usepackage{times}

\usepackage[numbers,sort,compress]{natbib}
\usepackage{multicol}

\usepackage{amsmath}
\usepackage{amssymb}
\usepackage{cases}
\usepackage{graphicx}
\usepackage{booktabs}

\usepackage{amsthm}

\theoremstyle{definition}

\usepackage[]{xcolor}
\usepackage{subcaption}
\usepackage{wrapfig}
\usepackage{multirow}
\usepackage{mathtools}
\usepackage{empheq}
\usepackage{color}
\usepackage{bbm}
\usepackage{xspace}
\let\labelindent\relax
\usepackage[shortlabels]{enumitem}

\usepackage{algorithmicx}
\usepackage{algorithm}
\usepackage[]{algpseudocode}

\usepackage{fontawesome}

\usepackage{soul}
\usepackage{comment}

\usepackage[]{url}

\usepackage[pdftex, colorlinks=true, linkcolor=black, citecolor = black, urlcolor = blue, filecolor=black , pagebackref=false, hypertexnames=false, bookmarks=false]{hyperref}
\usepackage{cleveref}
\crefname{problem}{Problem}{Problems}
\crefname{example}{Example}{Examples}
\crefname{section}{Sec.}{Secs.}
\Crefname{section}{Section}{Sections}
\Crefname{table}{Table}{Tables}
\crefname{table}{Table}{Tabs.}
\crefname{figure}{Fig.}{Figs.}
\crefname{algorithm}{Algorithm}{Algorithms}
\crefname{remark}{Remark}{Remarks}
\crefname{theorem}{Theorem}{Theorems}
\crefname{lemma}{Lemma}{Lemmas}
\crefname{corollary}{Corollary}{Corollaries}
\crefname{assumption}{Assumption}{Assumptions}
\usepackage[font=small]{caption}

\usepackage{siunitx}

%% file: include/macros.tex
\newcommand{\bmat}{\begin{bmatrix}}
\newcommand{\emat}{\end{bmatrix}}

\newcommand{\etal}{\emph{et~al.}\xspace}

\newcommand{\eg}{\emph{e.g.}\xspace}
\newcommand{\ie}{\emph{i.e.}\xspace}

\providecommand{\optional}[1]{{}}

\providecommand{\techreport}[1]{{}}

\newcommand{\map}[1][i]{\mathcal{M}_{#1}}
\newcommand{\M}[1]{{\mathbf #1}} %
\newcommand{\ve}[1]{{\mathbf #1}} %

\newcommand{\Tijhat}{\hat{\mathbf{T}}^i_j}
\newcommand{\R}{{\mathbb R}}

\makeatletter
\newcommand\notsotiny{\@setfontsize\notsotiny\@vipt\@viipt}
\makeatother

%% file: paper/abstract.tex
\begin{abstract}

Global localization is a fundamental capability required for long-term and drift-free robot navigation. 
However, current methods fail 
to relocalize when faced with significantly different viewpoints.
We present ROMAN (\underline{R}obust \underline{O}bject \underline{M}ap \underline{A}lignment A\underline{n}ywhere), a global localization method capable of localizing in challenging and diverse environments by creating and aligning maps of \mbox{\emph{open-set}} and \emph{view-invariant} objects. 
ROMAN formulates and solves a registration problem between object submaps using a unified graph-theoretic global data association approach with a novel incorporation of a gravity direction prior and object shape and semantic similarity.
This work's open-set object mapping and information-rich object association algorithm enables global localization, even in instances when maps are created from robots traveling in \emph{opposite} directions.
Through a set of challenging global localization experiments in indoor, urban, and unstructured/forested environments, we demonstrate that ROMAN achieves higher relative pose estimation accuracy than other image-based pose estimation methods or segment-based registration methods.
Additionally, we evaluate ROMAN as a loop closure module in large-scale multi-robot SLAM and show a 35\% improvement in trajectory estimation error compared to standard SLAM systems using visual features for loop closures.
Code and videos can be found at \url{https://acl.mit.edu/roman}.

\end{abstract}

%% file: paper/introduction.tex
\section{Introduction}

\emph{Global localization} \cite{yin2023survey} refers to the task of localizing a robot in a reference map produced in a prior mapping session or by another robot in real-time, \ie, inter-robot loop closures in collaborative SLAM \cite{lajoie2021towards}.
It is a cornerstone capability for drift-free navigation in GPS-denied scenarios.
In this paper, we consider global localization using \emph{object-} or \emph{segment-level} representations,\footnote[1]{We use \emph{object} and \emph{segment} interchangeably.} 
which have been shown by recent works~\cite{salas2013slampp,yu2022semanticloop,thomas24Sos, liu2024slideslam} to hold great promise in challenging domains that involve drastic changes in viewpoint, appearance, and lighting.

\begin{figure}[t]
    \centering
    \includegraphics[width=0.48\textwidth]{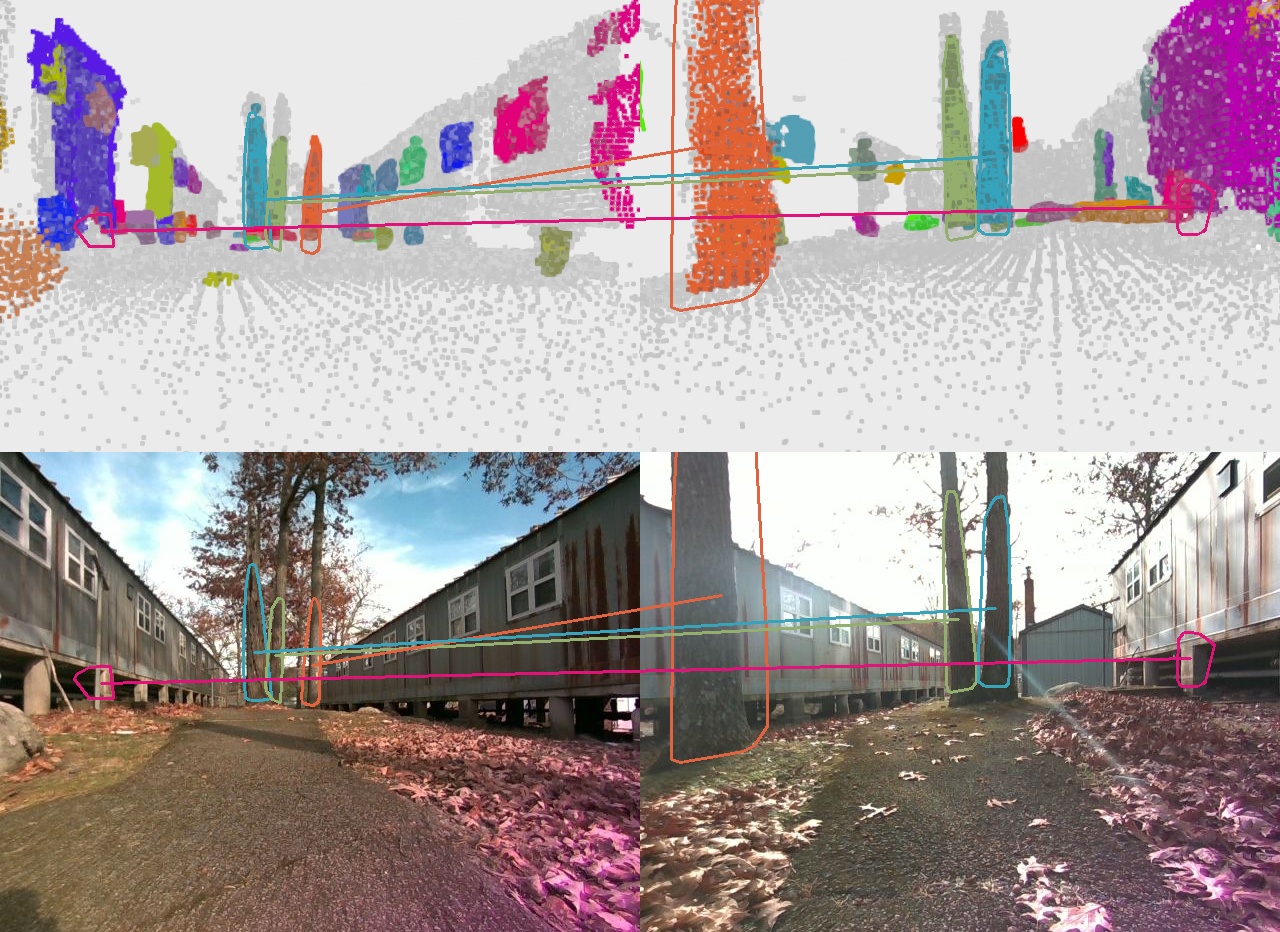}
    \caption{Pair of segment submaps matched by two robots traveling in \emph{opposite} directions in an off-road environment. Associated segments found by the proposed method are connected by lines and  projected onto the image plane. \textbf{(Top)} Each pair of associated segments is drawn with the same color. The remaining, unmatched segments are shown in random colors and all other background points are shown in gray. \textbf{(Bottom)} The same associated segments and their convex hulls are visualized in the original image observations. 
    Further visualization is shown in the supplementary video.
    }
    \label{fig:wp}
    \vspace{-0.4cm}
\end{figure}

At the heart of object-level localization is a \emph{global data association} problem, which requires finding {correspondences} between observed objects and existing ones in the map without an initial guess.
Earlier approaches such as \cite{dube2017segmatch,tinchev2018seeing,dube2020segmap,cramariuc2021semsegmap} rely on geometric verification based on RANSAC \cite{ransac}, which exhibits intractable computational complexity under high outlier regimes.
Recently, graph-theoretic approaches \cite{lusk2024clipper,mangelson2018pairwise,shi2021robin,forsgren2024group,yu2022semanticloop, dube2018incremental} have emerged as a powerful alternative that demonstrates superior accuracy and robustness when solving the correspondence problem.  
In particular, methods based on consistency graphs \cite{lusk2024clipper,mangelson2018pairwise,shi2021robin,forsgren2024group,dube2018incremental}
formulate a graph where nodes denote putative object correspondences and edges denote their geometric consistencies.
The data association problem is then solved by extracting large and densely connected subsets of nodes 
yielding the desired set of \emph{mutually consistent} correspondences.
While segment-based matching has become an established strategy for loop closures,
prior approaches were largely demonstrated in indoor/structured settings~\cite{sarkar2023sgaligner}, with limited object variations, or with accurate lidar sensing~\cite{dube2020segmap,dube2018incremental,li2021sa}.
In contrast, we focus on unseen environments (\ie, we do not make assumptions about the type of environment in which we operate), noisy segmentations, extreme viewpoint changes (\Cref{fig:wp}), and RGB-D only sensing.
Our key claim is that the proposed work is the only method that performs reliably in such extreme regimes and clearly outperforms state-of-the-art segment-based~\cite{lusk2024clipper,ransac,yang2020teaser} and visual-feature-based~\cite{sarlin2020superglue,leroy2024grounding} methods in global localization tasks.

Performance in these challenging scenarios is made possible 
by extending graph-theoretic data association to use information beyond mutual (pairwise) geometric consistency. 
We enhance the representational richness of association affinity metrics by developing a unified formulation that incorporates:
(i) \emph{open-set semantics}, extracted as semantically meaningful 3D segments \cite{ravi2024sam,zhao2023fast} with descriptors obtained from vision-language foundation model, CLIP \cite{clip};
(ii) \emph{segment-level geometric attributes}, such as the volume and 3D shapes of segments that provide additional discriminative power;
and (iii) an \emph{additional prior} about gravity direction that is readily available from onboard inertial sensors.

\textbf{Contributions.}
We present ROMAN (\underline{R}obust \underline{O}bject \underline{M}ap \underline{A}lignment A\underline{n}ywhere), a robust global localization method in challenging unseen environments. In detail, ROMAN consists of the following contributions:
\begin{enumerate}
    \item A graph-theoretic data association formulation with a novel method to incorporate 
    segment-level similarities computed using CLIP descriptors and geometric attributes based on shape and volume. 
    When gravity direction is known, a gravity-direction prior is also utilized.
    Our method implicitly guides the solver to correct 3D segment-to-segment associations in challenging regimes when object centroids alone are insufficient for identifying correct associations (\eg, due to repetitive geometric structures or scenes with few distinct objects)
    \item A pipeline for creating open-set 3D segment maps from a single onboard RGB-D camera, using FastSAM~\cite{zhao2023fast} for open-set image segmentation and CLIP \cite{clip} for computing open-set feature descriptors.
    These maps compactly summarize the detailed RGB-D point clouds into sparse and view-invariant representations consisting of segment locations and metric-semantic attributes, which enable efficient and robust global localization.
    \item Extensive experimental evaluation of the proposed method using real-world datasets (see \Cref{fig:wp}) that involve urban, off-road, and ground-aerial scenarios. 
    Our approach improves pose estimation accuracy by 45\% in challenging, opposite-view global localization problems.
    When using ROMAN rather than visual features for inter-robot loop closures in multi-robot SLAM, our method reduces the overall localization error by 8\% on large-scale collaborative SLAM problems involving 6-8 robots and by 35\% on a subset of particularly challenging sequences.
\end{enumerate}

%% file: paper/related_works.tex
\section{Related Works}

Object-based maps are lightweight environment representations that enable robots to match perceived objects with previously built object maps using object geometry or semantic labels as cues for object-to-object data association. 
Compared to conventional keypoints extracted from visual or lidar observations, \emph{object-} or \emph{segment-level} representations are more stable against sensor noise and viewpoint, lighting, or appearance changes, which often cause visual feature-based methods to fail~\cite{tian23iros-KimeraMultiExperiments}. Furthermore, these representations are lightweight and efficient to transmit, an important criterion for multi-robot systems. 
In this section, we review related methods for using object maps for global localization and SLAM.

\textbf{Object SLAM.} 
To incorporate discrete objects into SLAM, sparse maps of objects are described with geometric primitives such as points \cite{bowman2017probabilistic}, cuboids \cite{yang2019cubeslam} or quadrics \cite{nicholson2018quadricslam}. 
SLAM++~\cite{salas2013slampp} trains domain-specific object detectors for objects like tables and chairs. 
Choudhary~\etal~\cite{choudhary2014slam} use objects as landmarks for localization, providing a database of discovered objects. 
Lin~\etal~\cite{lin2021topology} showed that semantic descriptors can improve frame-to-frame object data association.
Recent works ~\cite{maggio2024clio, liu2024slideslam} further leverage \emph{open-set} semantics from pre-trained models.
Other methods~\cite{wang2024voom, zins2022oa} combine the use of coarse objects for high-level semantic information with fine features for high accuracy in spatial localization. 
Object-level mapping also conveniently handles dynamic parts of an environment which can be naturally described at an object level~\cite{tian2023object,schmid2024khronos}.

\textbf{Random sampling for object-based global localization.} Object-level place recognition may be performed by an initial coarse scene matching procedure (e.g., matching bag-of-words descriptors for scenes~\cite{hughes2024foundations}) but is commonly solved in conjunction with the object-to-object data association by attempting to associate objects and accepting localization estimates when object matches are good~\cite{gawel2018x, thomas24Sos}.
Object-to-object data association may be solved by sampling potential rotation and translation pairs between maps~\cite{liu2024slideslam} or object associations~\cite{dube2017segmatch, dube2020segmap, cramariuc2021semsegmap, tinchev2018seeing} using RANSAC~\cite{ransac}.
Random sampling methods often require significant computation for satisfactory results and the probability of finding correct inlier associations diminishes exponentially as the number of outliers grows~\cite{raguram2008comparative}. 

\textbf{Graph matching for object-based global localization.} 
Recently, graph-based methods have emerged as a fast and accurate alternative for object data association.
Objects are represented as nodes in a graph with graph edges encoding distance between objects~\cite{gawel2018x, yu2022semanticloop, wang2024goreloc}.
Data association can be performed by matching small, local target graphs with the prior map graph using graph-matching techniques.

\textbf{Maximal consistency for object-based global localization.}
Different from graph-matching methods, consistency graph algorithms use 
nodes to represent potential associations between two objects in different datasets, and edges to encode consistency between pairs of associations.
Data associations are found by selecting large subsets of mutually consistent nodes (associations), 
which can be formulated as either a maximum clique \cite{shi2021robin, mangelson2018pairwise, forsgren2024group, dube2018incremental} or densest subgraph \cite{lusk2024clipper} problem.
The work by Dub{\'e} \etal \cite{dube2018incremental} is one of the early works that performs global localization by finding maximum cliques of consistency graphs.
Ankenbauer \etal \cite{ankenbauer2023global} leverage graph-theoretic data association \cite{lusk2024clipper} as the back-end association solver to perform global localization in challenging outdoor scenarios. 
Matsuzaki~\etal~\cite{matsuzaki2024single} use semantic similarity between a camera image and a predicted image to evaluate pairwise consistency.
Thomas~\etal~\cite{thomas24Sos} 
use pre-trained, open-set foundation models for zero-shot segmentation in novel environments for open-set object map alignment.
Our method extends these prior works by incorporating object-to-object similarity and an additional pairwise association prior used to guide the optimization to correct associations.

\textbf{Inter-Robot Loop Closures for Collaborative SLAM.}
In the context of multi-robot collaborative SLAM (CSLAM),
our approach serves to detect \emph{inter-robot} loop closures that fuses individual robots' trajectories and maps.
State-of-the-art CSLAM systems \cite{tian2022kimera,schmuck2021covins,lajoie2023swarm,huang2021disco,chang2022lamp} commonly adopt a two-stage loop closure pipeline,
where a place recognition stage finds candidate loop closures by comparing global descriptors
and a geometric verification stage finds the relative pose by registering the two keyframes.
To improve loop closure robustness,
Mangelson~\etal~\cite{mangelson2018pairwise} proposes pairwise consistency maximization (PCM) which extracts inlier loop closures from candidate loop closures by solving a maximum clique problem.
Do~\etal~\cite{do2020robust} extends PCM \cite{mangelson2018pairwise} by incorporating loop closure confidence and weighted pairwise consistency.
Choudhary~\etal~\cite{choudhary2017distributed} performs inter-robot loop closure via object-level data association;
however, a database of 3D object templates is required.
Hydra-Multi~\cite{chang2023hydra} employs hierarchical inter-robot loop closure that includes places, objects, and visual features summarized in a scene graph.

%% file: paper/roman.tex
\section{ROMAN}
\label{sec:roman}

We now give an overview of the ROMAN global localization method.
The core idea behind this work is that small, local maps of objects near a robot give rich, global information about the robot's pose in a previously mapped area.
To leverage this information, ROMAN uses a mapping module to create object submaps and a robust data association module to associate objects in the robot's local map with objects seen by another robot or mapping session (see~\Cref{fig:pipeline}).

Our mapping pipeline begins with open-set image segmentation to extract initial observations of objects.
Then, object observations are aggregated into an abstract object map.
While we initially represent mapped objects with a dense point cloud, once the robot has moved on from an area, objects are abstracted to a single point and a feature descriptor, making our world representation communication- and storage-efficient.
A submap centered around a robot's pose and containing nearby sparse, abstract objects is then created and used for global localization.
Using local 3D segments, global localization can be achieved by matching objects in a local submap with objects from another robot or session. 
This is accomplished using our robust object data association method that leverages segment geometry, semantic information, and the direction of gravity to correctly associate objects.
Our view-invariant global localization formulation enables global localization even in cases when maps were created by robots traveling in opposite directions.
We first describe ROMAN's object data association method in~\Cref{sec:method-data-association} and then present our approach for creating open-set object maps in~\Cref{sec:method-mapping}.

\input{paper/figures/data_association}
\input{paper/figures/pipeline_diagram}
\subsection{Notation}
We use boldfaced lowercase and uppercase letters to denote vectors and matrices, respectively. 
We define $[n] = \{1, 2,\ldots,n\}$.
For any $n \in \mathbb{N}$ and $x_1, \dots, x_n \in \mathbb{R}$, we use 
$\text{GM}(x_1, \dots, x_n) \triangleq \left(\Pi_{i=1}^n x_i\right)^{\frac{1}{n}}$ to denote the geometric mean of $x_1, \ldots, x_n$, and $\text{GM}(\ve{x})$ to denote the geometric mean of the elements of the vector $\ve{x}$.
For any vectors $\ve{x}, \ve{y} \in \mathbb{R}^n$, their cosine similarity is denoted as $\text{cos\_sim}(\ve{x}, \ve{y}) \triangleq \frac{\left<\ve{x}, \ve{y}\right>}{\|\ve{x}\|_{2}\|\ve{y}\|_{2}}$.
We define the element-wise operation $\text{ratio}(\ve{x}, \ve{y}) \triangleq \min(\frac{\ve{x}}{\ve{y}}, \frac{\ve{y}}{\ve{x}})$, where $\min$ and $\frac{\ve{x}}{\ve{y}}$ are also performed element-wise. 
We use $\mathbf{T}^a_b \in \text{SE}(3)$ to denote the pose of frame $\mathcal{F}_b$ with respect to frame $\mathcal{F}_a$.

%% file: paper/figures/data_association.tex
\begin{figure}[!t]
    \centering
    \includegraphics[trim={3cm 1cm 3cm 1.5cm}, clip, width=\columnwidth] {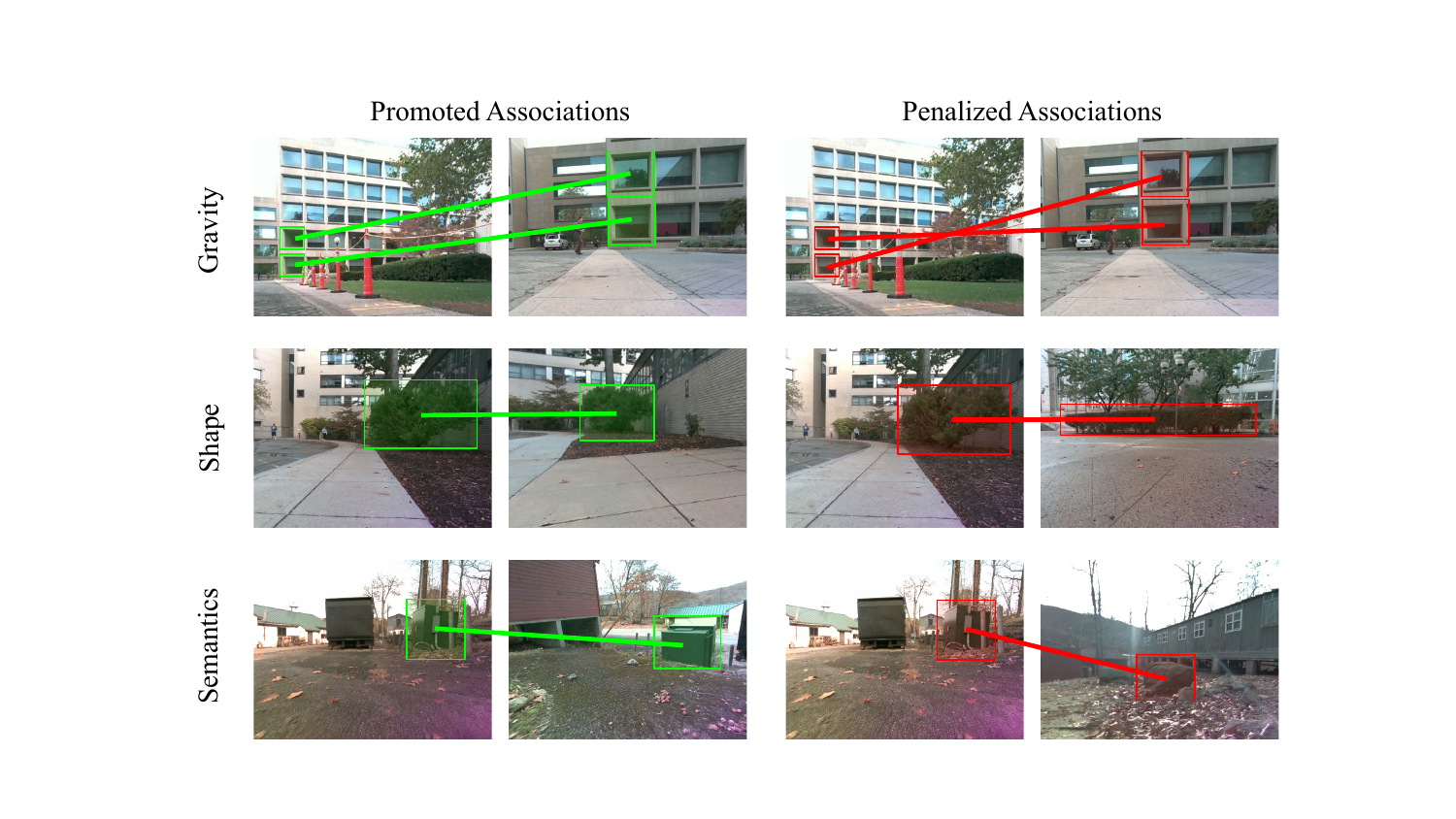}
    \caption{Visualization of improved affinity metrics. The gravity-based distance score, $s_\text{gravity}$ promotes pairs of associations that are consistent with the direction of gravity, while $s_\text{shape}$ and $s_\text{semantic}$ are used to encourage individual associations to be consistent in terms of geometric shape and semantics respectively.}
    \label{fig:data_association}
\end{figure}

%% file: paper/figures/pipeline_diagram.tex
\begin{figure*}[t]
    \centering
    \includegraphics[width=\textwidth]
    {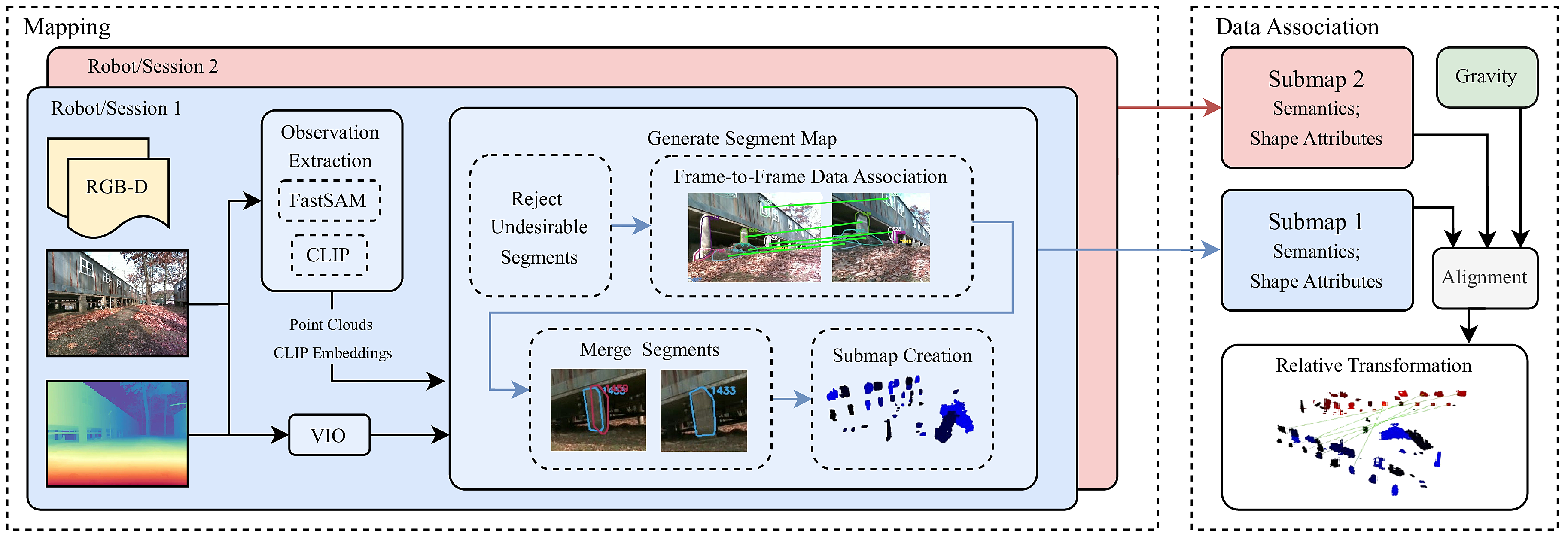}
    \caption{ROMAN employs a front-end mapping module to create maps of open-set objects, representing each object with its centroid and feature descriptor.
    Local collections of objects are grouped into submaps and used for global localization by matching objects between two submaps.
    Accurate data association is achieved using a graph-theoretic formulation which leverages object shape and semantic similarity and a gravity prior.
    }
    \label{fig:pipeline}
\end{figure*}

%% file: paper/data_association.tex
\section{Robust Object Data Association}
\label{sec:method-data-association}

While our data association method can be used for general point cloud registration, we focus on the problem of associating objects between two local object submaps for global localization.
We first detail submap alignment for global localization in~\Cref{sec:submap-align} then briefly review fundamentals in graph-theoretic data association in \Cref{sec:data-assoc-prelim} before describing the proposed affinity metrics for object association in \Cref{sec:da-gen,sec:da-sim,sec:da-grav}.

\subsection{Submap Alignment}
\label{sec:submap-align}

We consider a pair of submaps $\map[i]$ and $\map[j]$ which are associated with gravity-aligned poses $\M{T}^i_{\map[i]}$ and $\M{T}^j_{\map[j]}$.
Each submap $\map[i] = \{p_1, \ldots, p_{m_i}\}$ where $p_k$ is a 3D segment, represented by a 3D point in the gravity-aligned map frame $\mathcal{F}_{\map[i]}$ and a feature vector containing shape and semantic attributes (object feature descriptors are discussed in greater detail in~\Cref{sec:da-sim}).
We formulate global localization as the problem of estimating the transformation $\Tijhat$ which relates the two local frames $\mathcal{F}_i$ and $\mathcal{F}_j$.
To accomplish this, we attempt to associate objects in $\map[i]$ with objects in $\map[j]$.
After finding these associations, $\hat{\M{T}}^{\map[i]}_{\map[j]}$ can be computed using the closed-form Arun's method~\cite{arun1987least}, enabling the relation between frames $\mathcal{F}_i$ and $\mathcal{F}_j$ given that $\hat{\M{T}}^i_j = \M{T}^i_{\map[i]} \hat{\M{T}}^{\map[i]}_{\map[j]} \left(\M{T}^j_{\map[j]}\right)^{-1}$.
Thus, the core challenge in this global localization setup is to
correctly associate segments, a challenging task in the presence of uncertainty, outliers, and geometric ambiguity.
To this end, we construct a novel map-to-map object association method leveraging a graph-theoretic formulation incorporating the direction of gravity within maps and object shape and semantic attributes.

\subsection{Preliminaries: Graph-Theoretic Global Data Association}
\label{sec:data-assoc-prelim}
We follow the formulation used by CLIPPER~\cite{lusk2024clipper} by first constructing a consistency graph, $\mathcal{G}$, where each node in the graph is a putative association $a_p = (p_i, p_j)$ between a segment $p_i \in \map[i]$ and a segment $p_j \in \map[j]$.
Edges are created between nodes when associations are geometrically consistent with each other. 
Specifically, given two putative correspondences $a_p = (p_i, p_j)$ and $a_q = (q_i, q_j)$, CLIPPER declares that $a_p$ and $a_q$ are consistent if the distance between segment centroids in the same map is preserved, \ie, if $d(a_p, a_q) \triangleq \left\lvert\, \|\ve{c}(p_i) - \ve{c}(q_i)\| - \|\ve{c}(p_j) - \ve{c}(q_j)\| \,\right\rvert$ is less than a threshold $\epsilon$, where $\ve{c}(\cdot) \in \R^3$ is centroid position of a segment.
In this case, a weighted edge between $a_p$ and $a_q$ is created with weight
$s_a(a_p, a_q) \triangleq \exp\left(-\frac{1}{2}\frac{d(a_p, a_q)^2}{\sigma^2}\right)$.
Intuitively, $s_a(a_p, a_q) \in [0,1]$ scores the consistency between two associations, and $\epsilon$ and $\sigma$ are tuneable parameters expressing bounded noise in the segment point representation.

Given the consistency graph $\mathcal{G}$, a weighted affinity matrix $\M{M}$ is created where $\M{M}_{p,q} = s_a(a_p, a_q)$ and $\M{M}_{p,p} = 1$. 
CLIPPER determines inlier associations by (approximately) solving for the {densest} subset of consistent associations, formulated as the following optimization problem,
\begin{equation}
    \begin{split}
        \underset{\ve{u} \in \{0, 1\}^n}{\max}  &\frac{\ve{u}^\top \M{M} \ve{u}}{\ve{u}^\top \ve{u}}. \\
        \text{subject to} \quad &\ve{u}_p \ve{u}_q = 0 \; \text{if}\;  \M{M}_{p,q} = 0, \; \forall {p,q},
    \end{split}
    \label{eq:clipper}
\end{equation}
where $\ve{u}_p$ is 1 when association $a_p$ is accepted as an inlier and 0 otherwise.
In the following sections, we describe methods to improve affinity metrics. 
Given our construction of $\M{M}$, we then use CLIPPER's solver to find inlier associations $\ve{u}$.
See~\cite{lusk2024clipper} for more details.

\subsection{Improving affinity metrics: general strategies}
\label{sec:da-gen}

In its original form, the affinity matrix $\M{M}$ in \Cref{eq:clipper} relies solely on distance information between pairs of centroids.
However, when applied to segment maps,
unique challenges are introduced that are often not faced in other point registration problems (\eg, lidar point cloud registration), 
including dealing with greater noise in segment centroids (\eg, due to partial observation)
and few  inlier segments 
mapped in both $\map[i]$ and $\map[j]$, which
can lead to ambiguity when performing segment submap registration.
To address these problems, other works \cite{thomas24Sos, peterson2024motlee} have proposed pre-processing or post-processing methods that leverage additional information such as segment size and gravity direction to filter incorrect object associations or reject returned inlier associations if they result in an estimated $\Tijhat$ that is inconsistent with gravity.

In comparison to works that use prior information 
in pre-processing or post-processing steps which may discard valuable information,
ROMAN directly incorporates gravity and object similarity into the underlying optimization problem in~\Cref{eq:clipper}.
The key to our approach is to extend the original similarity metric to (i) use 
additional geometric (\eg, volume, spatial extent) and semantic (\eg, CLIP embeddings) attributes to 
disambiguate segments, 
and (ii) directly incorporate knowledge of the gravity direction (when available) to guide the data association solver. 

Consider the putative association $a_p = (p_i, p_j)$.
Intuitively, if objects $p_i$ and $p_j$ are dissimilar, then the association $a_p$ is less likely to be correct, which should be represented in the data association optimization formulation of~\Cref{eq:clipper}. 
Given a segment similarity score $s_o(a_p)$ comparing objects $p_i$ and $p_j$, 
~\cite{lusk2024clipper} and~\cite{leordeanu2005spectral} suggest setting the diagonal entries of $\M{M}$ to reflect object similarity information, \eg, by setting $\M{M}_{p,p} = s_o(a_p)$;
however, expanding the numerator of~\Cref{eq:clipper} shows that this approach has limited impact,
\begin{equation}
    \ve{u}^\top\M{M}\ve{u} = \Sigma_{p \in [n]} \left(
        \M{M}_{p, p} \ve{u}_p^2 +
        \Sigma_{q \in [n], q \neq p}\left(
            \M{M}_{p, q} \ve{u}_p \ve{u}_q
    \right)\right).
    \label{eq:clipper_objective_expanded}
\end{equation}
As the dimension of $\M{M}$ increases, 
the number of off-diagonal terms (pairwise association affinity terms) increases quadratically and will quickly dominate the overall objective function.
Alternatively,~\cite{do2020robust} and~\cite{yu2022semanticloop} propose multiplying the association affinity score by $s_o(\cdot)$ so that $\M{M}_{p,q} = s_a(a_p, a_q)s_o(a_p)s_o(a_q)$.
While this gives segment-to-segment similarity a significant role in the registration problem, 
the elements of $\M{M}$ are skewed to be much smaller resulting in many fewer accepted inlier associations.
To incorporate segment-to-segment similarity without significantly diminishing the magnitudes of the entries of $\M{M}$,
we instead propose using the \emph{geometric mean},
\begin{equation}
    \M{M}_{p,q} = 
    \text{GM}(s_a(a_p, a_q) , s_o(a_p), s_o(a_q)).
    \label{eq:geometric_mean}
\end{equation}

The use of geometric mean in merging scores of potentially different scales is well-studied in the field of operation research \cite{ROBERTS1994621}.
It was shown that, under reasonable assumptions, the geometric mean is the only averaging function that merges scores correctly~\cite{aczel1989possible,aczel1990determining}.
With this insight in mind, we incorporate additional information into the optimization problem~\eqref{eq:clipper} through careful designs of $s_a(\cdot, \cdot)$ and $s_o(\cdot)$, which will be explained in the subsequent subsections. 
An ablation study on fusion methods is presented in~\Cref{sec:exp-kmd-ablation}.

\subsection{Improving affinity metrics: incorporating metric-semantic segment attributes}
\label{sec:da-sim}

In this subsection, we design the segment-to-segment similarity score $s_o(\cdot)$ by comparing geometric and semantic attributes of the mapped segments (visualized in~\Cref{fig:data_association}). 
From the relatively dense point-cloud representation created for online mapping, a low-data shape descriptor and the averaged semantic feature descriptor are extracted for each 3D segment.
These descriptors are compared using a shape similarity scoring function $s_\text{shape}(\cdot)$ and a semantic similarity score $s_\text{semantic}(\cdot)$, which we present next.
The final segment-to-segment similarity score $s_o(\cdot)$ is set to be the geometric mean of those two scores. %

\paragraph{Semantic similarity metric}
To incorporate semantic information, we define the segment-to-segment semantic similarity score by taking the cosine similarity of their CLIP descriptors:  
$s_\text{semantic}(a_p) = \text{cos\_sim}(\text{CLIP}(p_i), \text{CLIP}(p_j))$.
We observe that the cosine similarity score of pairs of CLIP embeddings from images is usually higher than $0.7$, which does not allow semantic similarity to play a significant role in determining data associations in~\Cref{eq:clipper}.
We propose to rescale the cosine similarity score using hyperparameters $\phi_\text{min}$ and $\phi_\text{max}$, so that scores less than $\phi_\text{min}$ are set to 0, scores larger than $\phi_\text{max}$ are set to 1, and scores between $\phi_\text{min}$ and $\phi_\text{max}$ are scaled linearly so that they range from 0 to 1.

\paragraph{Shape similarity metric}
To incorporate segment shape attributes, 
we define a segment-to-segment shape similarity score:
\begin{equation}
    s_\text{shape}(a_p) = \text{GM}\left(
        \text{ratio}(\ve{f}(p_i), \ve{f}(p_j))
    \right),
\end{equation}
where $\ve{f}(p)$ returns a four-dimensional vector of the shape attributes of $p$ and is defined as follows. 
For each segment $p$, $\ve{f}_1(p)$ is the volume of the bounding box created from the point cloud of segment $p$, and 
$\ve{f}_2(p), \ve{f}_3(p)$, and $\ve{f}_4(p)$ denote the linearity, planarity, and scattering attributes of the 3D points computed via principle component analysis (PCA).
The interested reader is referred to \cite{weinmann2014semantic} for details.
The scoring function $s_\text{shape}(\cdot) \in [0, 1]$ allows direct feature element-to-element scale comparison.
Intuitively, if one element is much larger than the other, the score will be near $0$, while if the element is very similar in scale, $s_\text{shape}$ will be close to $1$.

\subsection{Improving affinity metrics: incorporating gravity prior}
\label{sec:da-grav}

We additionally address implicitly incorporating knowledge of the gravity direction in the global data association formulation. 
Due to the geometric-invariant formulation of~\Cref{eq:clipper}, the solver naturally considers registering object maps as a 6-DOF problem.
Often in robotics, an onboard IMU makes the direction of the gravity vector well-defined, so we are only interested in transformations with $x$, $y$, $z$, and yaw components.
Because the optimization variable of~\Cref{eq:clipper} is a set of associations rather than a set of transformations, it is not immediately clear how to leverage this information within the optimization problem, motivating the post-processing rejection step from~\cite{thomas24Sos}. 
In this work, we propose a method to leverage this extra knowledge \emph{within} the data association step by replacing $s_a(\cdot, \cdot)$ with a redesigned pairwise score metric, $s_\text{gravity}(\cdot, \cdot)$, to
guide the solver to select pairs of associations that are consistent with the direction of the gravity vector.
Specifically, we represent this prior knowledge of the gravity vector by decoupling computations in the $x$-$y$ plane and along the $z$ axis:
\begin{equation}
\label{eq:gravity-invariant}
    s_a(a_p, a_q) = \exp\left(-\frac{1}{2}\left(
        \frac{d^2_{xy}(a_p, a_q)}{\frac{2}{3}\sigma^2} + \frac{d^2_{z}(a_p, a_q)}{\frac{1}{3}\sigma^2}
    \right)\right),
\end{equation}
where 
\begin{align*}
d_{xy}(a_p, a_q) =& \left\lvert\, \|\ve{c}_{xy}(p_i) - \ve{c}_{xy}(q_i)\| 
- \|\ve{c}_{xy}(p_j) - \ve{c}_{xy}(q_j)\| \,\right\rvert \notag \\
d_{z}(a_p, a_q) =& \left\lvert\, (\ve{c}_{z}(p_i) - \ve{c}_{z}(q_i)) - (\ve{c}_{z}(p_j) - \ve{c}_{z}(q_j)) \,\right\rvert.
\end{align*}

In effect, this prohibits selecting pairs of associations where the vertical distances between objects within the same submap are dissimilar, as visualized in~\Cref{fig:data_association}. 
It is important to note that we use the \emph{difference} in the $z$-axis since we have directional information from the gravity vector while we only use \emph{distance} in the $x$-$y$ plane.
The directional information helps further disambiguate correspondence selection in scenarios where distance information is insufficient.

%% file: paper/mapping.tex
\section{Open-Set Object Mapping}
\label{sec:method-mapping}

This section describes ROMAN's approach to creating open-set object maps used for global localization in diverse environments. 
A map containing accurate and concise metric-semantic, object-level information is important for accurate object-based global localization.
However, creating such a map has historically been difficult due to the need for an object classifier. 
Using recent zero-shot open-set segmentation, object-level environment information can easily be extracted from each image, but aggregating this information is difficult due objects or groups of objects being segmented inconsistently between views, occluded object observations, and drift in robot odometry. 
To overcome these difficulties, we propose the following open-set object mapping pipeline, which is visualized in~\Cref{fig:pipeline}.

\subsection{Mapping}
\label{sec:mapping}

The inputs to ROMAN's mapping module consist of RGB-D images and robot pose estimates (\eg, provided by a visual-inertial odometry system). 
Per image object observations are made by segmenting a color image using FastSAM \cite{zhao2023fast} and applying a series of preprocessing steps to filter out undesirable segments. 
Distinct and stationary objects are most likely to be segmented consistently across different views, so our segment filtering aims to capture only such segments.
We use YOLO-V7 \cite{shi2022yolov} to reject segments containing people.
Additionally, we project segments into 3D using the depth image and remove large planar segments which are often large ground regions or non-distinct walls which cannot be represented well as an object.
Each of the remaining segments is fed into CLIP~\cite{clip} to compute a semantic descriptor.
Observations, made up of CLIP embeddings and 3D voxels, are then sent to a frame-to-frame data association and tracking module. 

Data association is performed between existing 3D segment tracks and incoming 3D observations by computing the grid-aligned voxel-based IOU between pairs of tracks and observations with 3D voxel overlap~\cite{schmid2024khronos}.
We use a global nearest neighbor approach~\cite{kuhn1955hungarian} to assign observations to existing object tracks and create new tracks for any unassociated observation.
Semantic descriptors of the associated segments are merged by taking a weighted average of descriptors of the existing segment and the incoming segment as in \cite{gu2024conceptgraphs}.
Because FastSAM may segment objects differently depending on the view, we create a merging mechanism to avoid duplications of the same object.
Specifically, 3D segments are merged based on high grid-aligned voxel IOU or when a projection of the two segments onto the image plane results in a high 2D IOU.
The result of our mapping pipeline is a set of open-set 3D objects with an abstractable representation. 
While performing mapping, objects are represented by dense voxels helping the frame-to-frame data association and object merging.
However, our global localization only uses a low-data representation of segments consisting of centroid position, shape attributes, and mean semantic embedding, which enables efficient map communication and storage.

\subsection{Submap Creation}
\label{sec:submaps}

As a robot travels, submaps are periodically created. 
After a robot's odometry estimate reaches a distance greater than $c_d$ from the previous submap pose, a new submap is instantiated. 
The new submap is assigned the current robot's pose with pitch and roll components removed using the IMU's gravity direction estimate, which ensures that objects are represented in a gravity-aligned frame for data association. 
All objects within a radius $r$ of the submap center are added, and objects continue to be added until the robot's distance from the submap center is greater than $r$.
The submap is then saved, after using a maximum submap size $N$ to remove objects (starting at objects farthest from the center) so that the submap size $m_i \leq N$ thus limiting submap alignment computation.
Finally, a newly created submap is fed to the global data association module and ROMAN attempts to align the current submap with previous submaps (e.g., from earlier in the run or from another robot or session).
Resulting $\Tijhat$ estimates from the submap object data association and alignment are used for global localization if the number of associated objects is greater than a threshold, $\tau$.

%% file: paper/experiment.tex
\section{Experiments}\label{sec:experiments}

In this section, we evaluate ROMAN in an extensive series of diverse, real-world experiments.
Our evaluation settings consist of urban domains from the large-scale Kimera-Multi datasets \cite{tian23iros-KimeraMultiExperiments}, 
off-road domains in an unstructured, natural environment,
and ground-aerial localization in a manually constructed, cluttered indoor environment.
Experimental results demonstrate that ROMAN achieves superior performance compared to existing baseline methods, 
obtaining up to 45\% improvement in relative pose estimation accuracy in opposite directions and 35\% improvement in final trajectory estimation error 
in a subset of particularly challenging sequences from the Kimera-Multi datasets. 
The experiments were run on a laptop with a
4090 Mobile GPU and a 32-thread i9 CPU.

\subsection{Experimental Setup}
\label{sec:exp-setup}

\textbf{Baselines.} We compare the alignment performance of \texttt{ROMAN} 
against the following baselines.
\texttt{RANSAC-100K} and \texttt{RANSAC-1M} apply RANSAC~\cite{ransac}, as implemented in~\cite{zhou2018open3d}, on segment centroids with a max iteration count of 100,000 and 1 million respectively.
\texttt{CLIPPER} runs standard CLIPPER~\cite{lusk2024clipper} on segment centroids, and \texttt{CLIPPER / Prune} prunes initial putative associations using semantic and shape attributes and rejects incorrect registration results using gravity information (so it has access to similar information as the proposed method).
\texttt{TEASER++ / Prune} runs the robust registration of~\cite{yang2020teaser} using the same pruning mechanism as \texttt{CLIPPER / Prune}.
\texttt{Binary Top-K}, which mimics the association method of SegMap~\cite{dube2020segmap}, takes the top-k most similar segments (in terms of the semantic and shape descriptors) and constructs a binary affinity matrix that we use for finding associations with solver from~\cite{lusk2024clipper}.
We also compare against recent image-based pose estimation methods.
\texttt{MASt3R} and \texttt{MASt3R} (GT Scale) use the learned 3D reconstruction model of~\cite{leroy2024grounding} to estimate relative camera poses with the model's estimated translation scale and the ground truth translation scale respectively.
\texttt{SuperGlue} (GT Scale)  similarly estimates relative camera poses using~\cite{sarlin2020superglue} to match SuperPoint features~\cite{detone2018superpoint}.
Additionally, we incorporate ROMAN as a loop closure detection module in single-robot and multi-robot SLAM and compare against \texttt{KM} (Kimera-Multi ~\cite{tian2022kimera}) and \texttt{ORB3} (ORB-SLAM3~\cite{campos2021orb}) which both use BoW descriptors of ORB features for loop closures. 

\textbf{Performance metrics.} We use the following metrics for comparing segment-based place recognition, submap alignment (equivalent to relative pose estimation for image-based methods),and full SLAM results.
For place recognition, each algorithm is given a query submap and a database composed of submaps from every other robot run.
Submap registration is performed on the query submap and every submap from the database.
The database submap with the highest number of associations is returned and success is achieved if the query and returned submaps overlap.
We vary the threshold on number of required object association $\tau$ to generate precision-recall curves, and following~\cite{zaffar2021vpr}, precision-recall area under the curve (AUC) is reported.

To evaluate alignment success rate, an algorithm is given a pair of submaps whose center poses are within \SI{10}{m} of each other.
We evaluate the image-based methods by giving an algorithm the two images corresponding to the two submap center poses.
To avoid giving segment-based methods an unfair advantage, we do not include submaps whose camera fields of view (FOVs) do not overlap.
Following~\cite{sarlin2022lamar}, alignment (\ie, pose estimation) success is determined when the transformation error is less than \SI{1}{m} and \SI{5}{deg}, with respect to ground truth. 

Full SLAM results are evaluated using root mean squared (RMS) absolute trajectory error (ATE) between the registered estimated and ground truth multi-robot trajectories.
We use open-source evo~\cite{grupp2017evo} to compute ATE.

\textbf{Parameters.} For global localization, we use the parameter values outlined in Table \ref{table:params}.
We additionally include results for two larger variants of our work: \texttt{ROMAN-L}, which uses $r = 25, N = 60$, and \texttt{ROMAN-XL}, which uses $r = 30, N=80$.
In pose graph optimization, we use odometry covariances with uncorrelated rotation and translation noise parameters.
We use standard deviations of 0.1 m and 0.5 deg for sparse odometry and  1.0 m and 2.0 deg for loop closures.

\input{tables/params}

\subsection{MIT Campus Global Localization}
\label{sec:exp-kmd-global-loc}

We first evaluate ROMAN's map alignment using the outdoor Kimera-Multi Dataset~\cite{tian23iros-KimeraMultiExperiments} recorded on MIT campus.
Each robot creates a set of submaps using Kimera-VIO~\cite{rosinol2020kimera} for odometry and our ROMAN mapping pipeline.
We use these submaps to evaluate segment-based place recognition and submap alignment for global localization, as described in~\Cref{sec:exp-setup}.
We evaluate methods on all multi-robot submap pairs from this dataset that are within \SI{10}{m} of each other and whose corresponding camera FOVs overlap.
In~\Cref{tab:kmd-glob-loc}, we show place recognition and submap alignment results.
To highlight performance across different viewpoints, we bin the alignment tests into three different ground-truth relative heading groups: 
$\theta \leq$ \SI{60}{deg} (same direction), \SI{60}{deg} $< \theta \leq$ \SI{120}{deg} (perpendicular), and $\theta >$ \SI{120}{deg}.
When the heading difference is small, alignment is comparatively easier.
Aligning submaps from opposite views or from paths that cross perpendicularly, presents the hardest cases for global localization.

~\Cref{tab:kmd-glob-loc} shows that the \texttt{ROMAN} outperforms other segment-based methods in terms of place recognition and alignment success rate in all heading intervals while operating at a similar runtime. 
In opposite directions, \texttt{ROMAN} achieves a pose estimation success rate 75\% higher than the next-best segment-based method, \texttt{CLIPPER/Prune}.
Compared to image-based methods, the ROMAN variant with more objects, \texttt{ROMAN-XL} outperforms the next-best method, \texttt{MASt3R} (which is given ground truth scale), in every case except for in similar direction scenarios, all while running 10 times faster. 
In particular, \texttt{ROMAN-XL} achieves a pose estimation success rate in opposite directions that is 45\% better than $\texttt{MASt3R}$ (GT Scale) and 31\% better when averaged across the different headings.

In terms of communication and submap storage size, each object includes a 3D centroid, a four-dimensional shape descriptor and a 768-dimensional semantic descriptor.
With each submap consisting of at most $N = 40$ objects, a submap packet size is strictly less than 250 KB. 
For a trajectory of length \SI{1}{km}, the entire map could be represented with less than 25 MB of data.

\input{tables/kmd}

\subsection{Loop Closures in Visual SLAM}
\label{sec:exp-kmd-rpgo}

We integrate ROMAN as a loop closure detection module for single and multi-robot pose-graph SLAM and compare the trajectory estimation results here and in~\Cref{sec:exp-offroad-align}.
We use Kimera-VIO~\cite{rosinol2020kimera} for front-end odometry when creating initial ROMAN submaps.
Then, we attempt to register each new submap with all existing submaps from the ego robot and other robots. 
Loop closures are reported when the number of associations found is at least $\tau$.
Then, sparsified Kimera-VIO odometry and ROMAN loop closures are fed into the robust pose graph optimization of Kimera-Multi~\cite{tian2022kimera} to estimate multi-robot trajectories.
Root-mean-squared (RMS) absolute trajectory errors (ATE) in the tunnel, hybrid, and outdoor Kimera-Multi datasets are reported in \Cref{table:rpgo}.
We compare SLAM with ROMAN loop closures against a centralized Kimera-Multi (\texttt{KM})~\cite{tian2022kimera} and multi-session ORB-SLAM3 (\texttt{ORB3})~\cite{campos2021orb}.
Note that in the single-robot case, the baselines are essentially a single-robot version of Kimera and single-robot ORB-SLAM3, where a deeper comparison was made in~\cite{abate2023kimera2}.
Similar to ~\cite{abate2023kimera2}, we found that ORB-SLAM3 fails to find reasonable trajectory estimates in some robot configurations, and this is represented with a dash in~\Cref{table:rpgo}.

Estimation errors show that, on average, in easier single-robot tunnel runs, ROMAN loop closures result in lower trajectory errors than ORB-SLAM3 and errors comparable to Kimera-Multi.
The full, large-scale multi-robot runs show that ROMAN's ability to detect loop closures in challenging visual scenarios results in moderate gains compared to Kimera-Multi's trajectory errors.
Improvement is somewhat limited due to the high-connectivity of robot paths and the fact that most robot trajectory overlap occurs when robots are traveling in the same direction, which are loop closure opportunities in which visual-feature-based methods already perform well.
However, when SLAM results are compared on a subset of robot trajectories that contain difficult instances for visual loop closures (\eg, perpendicular path crossing and scenes with high visual aliasing), results show that ROMAN has a significantly lower ATE in these challenging scenarios.
The trend is that as loop closure scenarios become increasingly difficult, ROMAN demonstrates more significant improvements over state-of-the-art methods.

\input{tables/kmd_rpgo}

\subsection{Loop Closures in Off-Road Environment}
\label{sec:exp-offroad-align}

\input{paper/figures/wp_rpgo}

We further evaluate the proposed method's ability to register segment maps in an outdoor, off-road environment with high visual ambiguity (\Cref{fig:wp}).
In this experiment, data is recorded on a Clearpath Jackal using Intel RealSense D455 to capture RGB-D images and Kimera-VIO~\cite{rosinol2020kimera} is used for odometry. 
The robot is teleoperated across four runs, following similar trajectories but with different runs traversing the same area while traveling in different directions.
We run the ROMAN pipeline on three different pairs of robot trajectories.
We compare \texttt{ROMAN} to \texttt{KM} loop closures in ~\cref{fig:wp_rpgo}.
The three pairs consist of an easy, medium, and hard case.
The easy case involves two robots that traverse the same loop in the same direction (with one robot that leaves the loop and later returns).
In the medium difficulty case, the robots travel in opposite directions except for a short section in the middle where both robots briefly view the scene from the same direction.
Finally, in the difficult case, robots travel in a large loop in opposite directions.
While ground-truth pose is not available for this data,
\Cref{fig:wp_rpgo} qualitatively shows that \texttt{ROMAN} successfully detects loop closures in all three cases.
More importantly, \texttt{ROMAN} successfully closes loops in opposite-direction traversals, while loop closures from \texttt{KM} only work reliably in same-direction traversals and fail to find any loop closures in the hard case.

\subsection{Ground-Aerial Cross-View Localization}
\label{sec:exp-aerial}

We also evaluate ROMAN's robustness to view changes by conducting indoor localization experiments where segment maps created from ground views are aligned with segment maps created from aerial views. 
Snapshots of the setup from both views are shown in \cref{fig:highbay_setup}.
We test object map alignment on 20 ground-aerial pairs of traverses through the environment, and
report alignment success rate in~\Cref{table:highbay_result}.
We show that ROMAN maintains an advantage over other baselines, demonstrating its global localization capability in a small-scale aerial-ground cross-view localization demonstration.

\input{paper/figures/highbay_setup.tex}

\input{tables/aerial.tex}

%% file: tables/params.tex
\begin{table}[h!]
\scriptsize
\centering
\caption{Parameters}

\centering
\begin{tabular}{c c c}
\toprule
\textbf{Parameter} & \textbf{Value} & \textbf{Description} \\
\midrule
$\sigma$ / $\epsilon$ & \SI{0.4}{m} / \SI{0.6}{m} & Pairwise consistency noise parameters \\
$r$ / $c_d$ & \SI{15}{m} / \SI{10}{m} & Submap radius and spacing distance \\
$\phi_\text{min}$ / $\phi_\text{max}$ & 0.85 / 0.95 & Cosine similarity scaling values \\
$\tau$ / $N$ & 4 / 40 & Association threshold and max submap size \\
\bottomrule
\end{tabular}

\label{table:params}
\end{table}

%% file: tables/kmd.tex
\begin{table}[t]
    \scriptsize
    \centering
    \caption{Kimera-Multi Outdoor Global Localization Results}

    \setlength{\tabcolsep}{1.5pt}
    \begin{tabular}{c l c c c c c c}
        \toprule
        & \multirow{3}{*}{\textbf{Method}} 
        & {\textbf{Place}} 
        & \multicolumn{4}{c}{\textbf{Pose Estimation Success Rate}} 
        & \multirow{2}{*}{\textbf{Runtime}} \\
        & & \textbf{Recognition} & \multicolumn{4}{c}{\textbf{($\leq$ 5$^\circ$, 1 m error)}} & \\
        & & \textbf{(AUC)} & \textbf{0--60$^\circ$} & \textbf{60--120$^\circ$} & \textbf{120--180$^\circ$} & \textbf{Mean} & \textbf{(ms)} \\
        \midrule
        
        \multirow{6}{*}{\rotatebox{90}{\textbf{Segment-Based}}} 
        & \texttt{RANSAC-100K}      & 0.106 & 0.141 & 0.000 & 0.000 & 0.047 & 75.9 \\
        & \texttt{RANSAC-1M}        & 0.160 & 0.341 & 0.065 & 0.054 & 0.153 & 488 \\
        & \texttt{CLIPPER}          & 0.145 & 0.235 & 0.043 & 0.000 & 0.093 & 48.2 \\
        & \texttt{CLIPPER/Prune}    & 0.531 & 0.429 & 0.109 & 0.108 & 0.215 & \underline{23.9} \\
        & \texttt{TEASER++/Prune}   & 0.426 & 0.441 & 0.125 & 0.083 & 0.216 & 498.6 \\
        & \texttt{Binary Top-K}     & 0.307 & 0.377 & 0.130 & 0.054 & 0.187 & \textbf{21.3}  \\
        \midrule

        \multirow{3}{*}{\rotatebox{90}{\textbf{Visual}}} 
        & \texttt{SuperGlue} (GT Scale) & $-$ & 0.685 & 0.043 & 0.000 & 0.243 & 87.4 \\
        & \texttt{MASt3R} (GT Scale)  & $-$ & \textbf{0.775} & 0.152 & 0.297 & 0.408 & 2950 \\

        & \texttt{MASt3R}               
            & $-$ & 0.211 & 0.043 & 0.000 & 0.085 & 2950 \\
        \midrule

        \multirow{3}{*}{\rotatebox{90}{\textbf{Ours}}} 
        & \texttt{ROMAN}     & 0.552 & 0.521 & 0.152 & 0.189 & 0.287 & {28.9} \\
        & \texttt{ROMAN-L}  & \textbf{0.663} & 0.723 & \underline{0.370} & \textbf{0.432} & \underline{0.508} & 92.9 \\
        & \texttt{ROMAN-XL}  & \underline{0.654} & \underline{0.745} & \textbf{0.457} & \underline{0.405} & \textbf{0.536} & 213 \\
        \bottomrule
        \vspace{-.9cm}
    \end{tabular}
    \label{tab:kmd-glob-loc}
\end{table}

%% file: tables/kmd_rpgo.tex
\begin{table}[t]
    \scriptsize
    \centering
    \caption{Kimera-Multi Data~\cite{tian23iros-KimeraMultiExperiments} SLAM Comparison Against Various Loop Closure Methods (RMS ATE m)}

    \begin{tabular}{ l c c c c c}
        \toprule
        \multirow{2}[2]{*}{\textbf{Dataset}} & \textbf{Num.} & \textbf{Total} & \multirow{2}[2]{*}{\texttt{ORB3}~\cite{campos2021orb}} & \multirow{2}[2]{*}{\texttt{KM}~\cite{tian2022kimera}} & \multirow{2}[2]{*}{\texttt{ROMAN}} \\
         & \textbf{Robots} & \textbf{Dist. (m)} \\
        \midrule

        \multicolumn{6}{c}{\textbf{Easy: Single Robot Tunnels}} \\
        \midrule
        Tunnel 0 & 1 & 635 & \textbf{2.08} & 4.20 & 4.16 \\
        Tunnel 1 & 1 & 780 & 26.19 & \textbf{1.61} & 2.15 \\
        Tunnel 2 & 1 & 854 & 9.53 & \textbf{5.29} & 6.12 \\
        Tunnel 3 & 1 & 845 & 16.61 & 5.29 & \textbf{3.90} \\
        Mean & & & 13.60 & 4.10 & \textbf{4.08} \\
        \midrule

        \multicolumn{6}{c}{\textbf{Medium: Full Multi-Robot Datasets}} \\
        \midrule
        Tunnel All & 8 & 6753 & -- & 4.38 & \textbf{4.20} \\
        Hybrid All & 8 & 7785 & -- & 5.83 & \textbf{5.12} \\
        Outdoor All & 6 & 6044 & -- & 9.38 & \textbf{8.77} \\
        Mean & & & -- & 6.53 & \textbf{6.03} \\
        \midrule

        \multicolumn{6}{c}{\textbf{Difficult: Challenging Multi-Robot Combinations}} \\
        \midrule
        Hybrid 1, 2, 3 & 3 & 3551 & -- & 10.34 & \textbf{6.91} \\
        Hybrid 4, 5 & 2 & 1896 & 28.09 & 6.11 & \textbf{2.80} \\
        Outdoor 1, 2 & 2 & 2011 & 11.93 & 10.12 & \textbf{7.67} \\
        Mean & & & -- & 8.86 & \textbf{5.79} \\
        
        \bottomrule
    \end{tabular}
    \label{table:rpgo}
\end{table}

%% file: paper/figures/wp_rpgo.tex
\begin{figure}[t!]
    \centering
    \vspace{-.2cm}
    \includegraphics[width=0.95\columnwidth, trim={1.0cm 1.3cm 1.0cm 1.0cm}, clip]{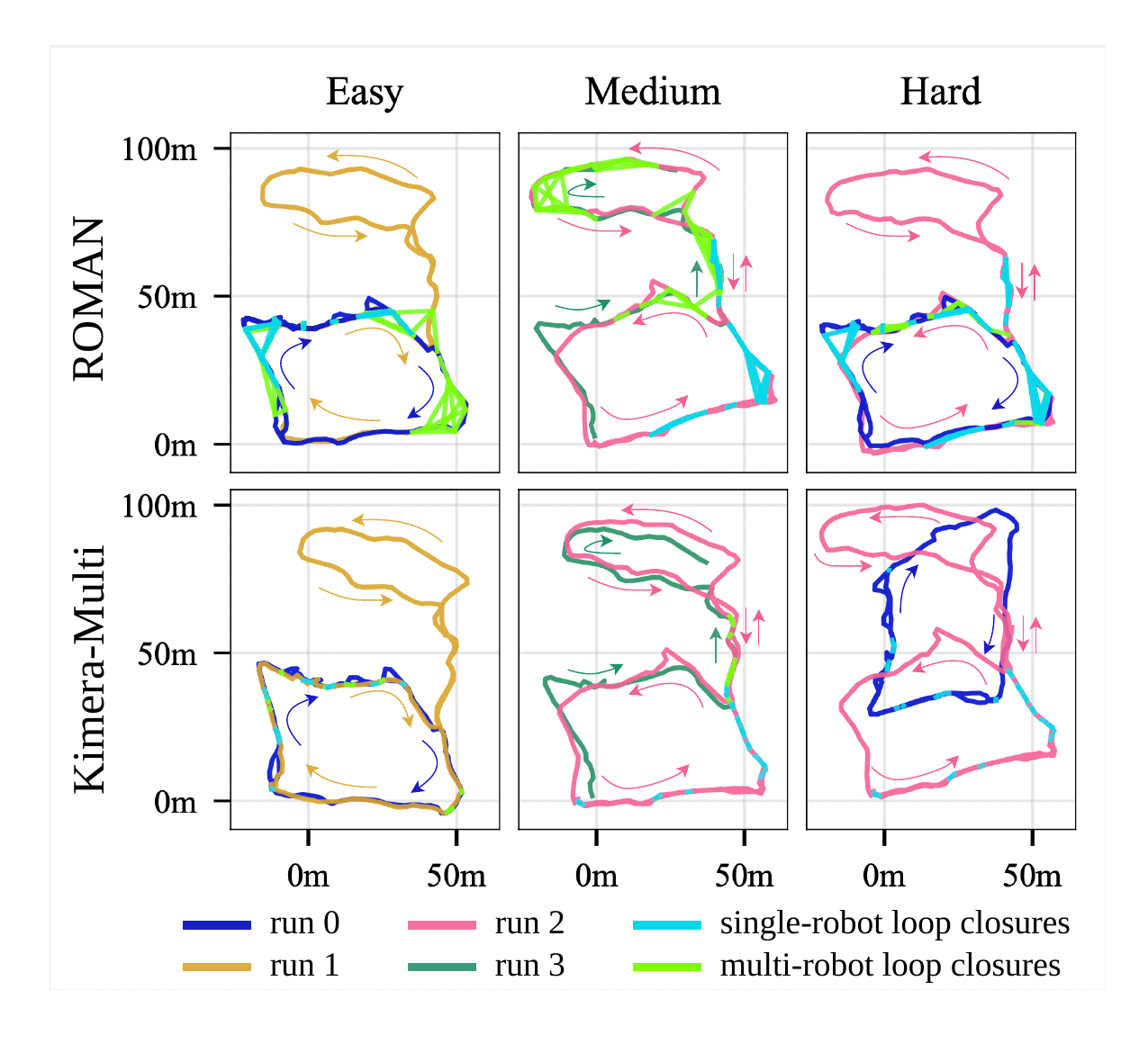}
    \caption{Off-road qualitative pose graph trajectory estimate. 
    Easy, medium, and hard cases comparing using \texttt{ROMAN} and \texttt{KM} for loop closures. 
    Different combinations were paired together to make easy, medium, and hard cases.
    In the easy case, robots travel in the same direction; in the medium case, the two runs go in opposite directions except for the small connecting neck; and in the hard case, robots only cross paths going in opposite directions.
    Only \texttt{ROMAN} successfully finds loop closures between robots running in opposite directions.}
    \label{fig:wp_rpgo}
\end{figure}

%% file: paper/figures/highbay_setup.tex
\begin{figure}[t]
    \centering
    \includegraphics[width=\columnwidth, height=2.5cm] {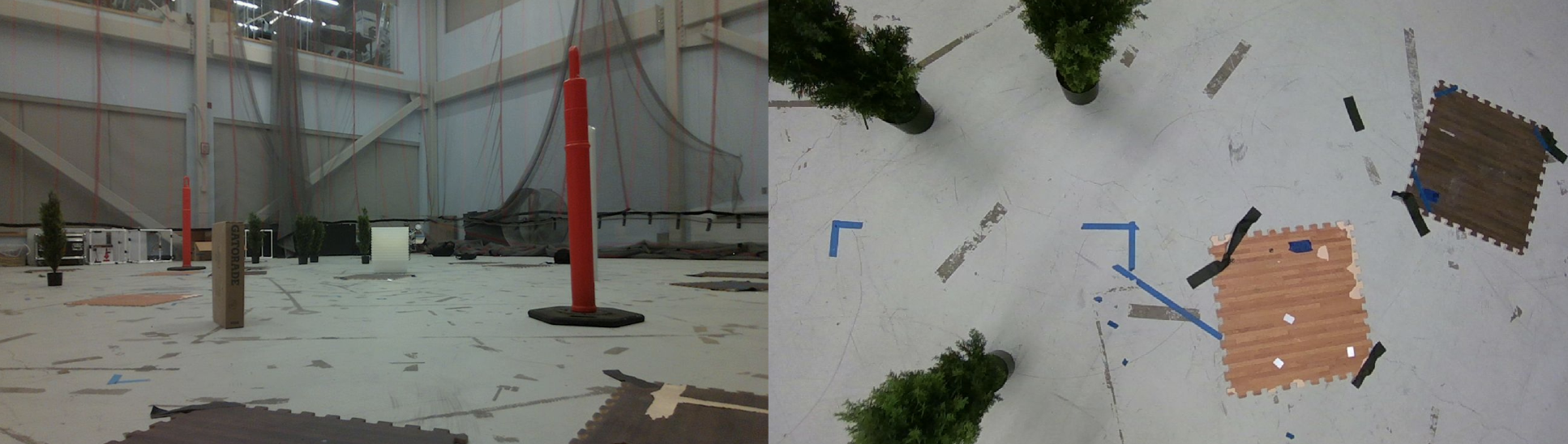}
    \caption{Environmental setup used in the ground-aerial cross-view localization experiment as seen from both ground view (left) and aerial view (right).}
    \label{fig:highbay_setup}
\end{figure}

%% file: tables/aerial.tex
\begin{table}[ht]
    \centering
    \caption{Ground-Aerial Cross-View Localization Results}
    \begin{tabular}{lclc}
        \toprule
        \textbf{Method} & \textbf{Recall} & \textbf{Method} & \textbf{Recall} \\
        \cmidrule(lr){1-2} \cmidrule(lr){3-4}
        \texttt{RANSAC-100K} & 0.25 & \texttt{RANSAC-1M} & 0.45 \\
        \texttt{CLIPPER} & 0.3  & \texttt{CLIPPER/Prune} & 0.35  \\
        \texttt{TEASER++/Prune} & 0.55 & \texttt{ROMAN} & 0.60 \\
        \bottomrule
    \end{tabular}
\label{table:highbay_result}
\end{table}

%% file: paper/ablations.tex
\subsection{Ablation Study}
\label{sec:exp-kmd-ablation}

Finally, we perform an extensive set of ablation studies examining the contribution of different affinity metric improvements, fusion methods, and other algorithmic elements.

\textbf{Fusion methods.}
Here, we compare different methods for fusing object similarity scores $s_o$ with pairwise scores $s_a$ in \Cref{tab:ablations}.
We investigate fusing scores with geometric mean (\texttt{ROMAN}), product ~\cite{do2020robust,yu2022semanticloop}, arithmetic mean, and 
setting the diagonal elements of the affinity matrix $M_{pp} = \text{GM}(s_o(a_p) s_o(a_q))$~\cite{leordeanu2005spectral,lusk2024clipper}.
\Cref{tab:ablations} shows that fusing scores using the geometric mean results in a much higher alignment success rate compared to other fusion methods.
Intuitively, fusing scores using the arithmetic mean has fewer zeroed-out elements of the affinity matrix which results in the optimization problem becoming less well-constrained.
Fusing via the product of scores improves the alignment success, but tends to over-penalize, since in this case, including $s_o$ can only lower the overall similarity score.
Changing only the diagonal elements also improves over standard CLIPPER~\cite{lusk2024clipper}, but is limited in impact as described in~\Cref{sec:da-gen}.

\textbf{Affinity component contributions.}
In~\Cref{tab:ablations}, we additionally examine the effect of using ROMAN for map alignment while excluding the following individual affinity metric components: the gravity-guided pairwise score $s_a$, the shape similarity score $s_\text{shape}$, and the semantic similarity score $s_\text{semantic}$. 
While each component helps ROMAN achieve higher alignment success, the gravity prior makes the most significant difference and the semantic similarity score makes the least.
However, in terms of place recognition, semantics makes the largest difference.

\textbf{Robustness to segmentation errors.}
As a small experiment, we change the input image size from 256 (the default value for which ROMAN is tuned) to obtain degraded segmentation (128) and over-segmentation (512).
On average, FastSAM~\cite{zhao2023fast} returns 4.0 segments at image size 128, 11.3 at 256, and 18.7 at 512.
As shown in~\Cref{tab:ablations}, in the case of over-segmentation, we report only 12\% performance decrease in terms of mean pose estimation success rate. 
With severe under-segmentation, ROMAN achieves 0.184 mean success, which is slightly lower than the best segment-based baselines shown in~\Cref{tab:kmd-glob-loc}, however some of the effects of under-segmentation could be mitigated by including segments in a larger radius $r$.

\textbf{Robustness to dynamic objects.}
The ROMAN pipeline deliberately filters out pedestrians, and the robust data association effectively rejects other dynamic objects.
To demonstrate the effect of dynamic objects, we disable the pedestrian filter and report that ROMAN achieves a mean alignment success rate of 0.251, which is still better than other segment-based baselines in~\Cref{tab:kmd-glob-loc}.

\textbf{Hyperparameter sensitivity.}
\Cref{tab:kmd-glob-loc} shows the effect of varying ROMAN submap sizes,  controlled by $N$ (maximum submap size) and $r$ (submap radius).
We vary $N$ from the default value 40 to 80 with $r$ increasing from \SI{15}{m} to \SI{30}{m} correspondingly.
The results show that these two submap size parameters can be effectively altered to achieve a trade-off between alignment success rate and runtime.
An ablation over the segment noise parameters $\sigma$ and $\epsilon$ are recorded in~\Cref{tab:ablations}.
We note that the lowest mean recall over all pairs is still higher than the mean recall of any other segment-based method in~\Cref{tab:kmd-glob-loc}.

\begin{table}[t]
    \scriptsize
    \centering
    \caption{Ablations Results}

    \setlength{\tabcolsep}{2pt}
    \begin{tabular}{c l c c c c c}
        \toprule
        & \multirow{3}{*}{\textbf{Ablations}} 
        & {\textbf{Place}} 
        & \multicolumn{4}{c}{\textbf{Pose Estimation Success Rate}} \\
        & & \textbf{Recognition} & \multicolumn{4}{c}{\textbf{($\leq$ 5$^\circ$, 1 m error)}} \\
        & & \textbf{(AUC)} & \textbf{0--60$^\circ$} & \textbf{60--120$^\circ$} & \textbf{120--180$^\circ$} & \textbf{Mean}\\
        \midrule
        
        & \texttt{ROMAN}     & 0.552 & 0.521 & 0.152 & 0.189 & 0.287 \\
        \midrule
        
        \multirow{3}{*}{\rotatebox{90}{\textbf{Exclude}}} 
        & Gravity   & 0.522 & 0.474 & 0.109 & 0.081 & 0.221  \\
        & Semantics     & 0.497 & 0.500 & 0.146 & 0.167 & 0.271  \\
        & Shape      & 0.530 & 0.532 & 0.152 & 0.108 & 0.264 \\
        \midrule
        
        \multirow{3}{*}{\rotatebox{90}{\textbf{Fusion}}} 
        & Arithmetic Mean     & 0.517 & 0.199 & 0.000 & 0.054 & 0.085 \\
        & Product      & 0.505 & 0.388 & 0.087 & 0.027 & 0.167 \\
        & Diagonal         & 0.336 & 0.388 & 0.109 & 0.027 & 0.175 \\

        \midrule
        \multirow{3}{*}{\rotatebox{90}{$(\sigma, \epsilon)$}} 
        & $(0.2, 0.3)$   & 0.504 & 0.433 & 0.229 & 0.028 & 0.230  \\
        & $(0.6, 0.9)$        & 0.543 & 0.522 & 0.104 & 0.139 & 0.255  \\
        & $(0.8, 1.2)$    & 0.535 & 0.548 & 0.125 & 0.194 & 0.289  \\
        \midrule
        
        \multirow{2}{*}[0.6ex]{%
        \rotatebox{90}{%
        \scriptsize
        \shortstack{\textbf{Image}\\\textbf{Size}}%
        }}
        & $128\times 128$     & 0.378 & 0.296 & 0.149 & 0.108 & 0.184 \\
        & $512 \times 512$      & 0.525 & 0.535 & 0.085 & 0.135 & 0.252 \\
        \bottomrule
    \end{tabular}
    \label{tab:ablations}
\end{table}

\textbf{Scalability.}
Our mapping pipeline runs at \SI{9.6}{Hz} when computing CLIP~\cite{clip} embeddings and at \SI{17.9}{Hz} without running CLIP on the outdoor Kimera-Multi Dataset~\cite{tian23iros-KimeraMultiExperiments}
As shown in~\Cref{tab:ablations}, alignment success rate only drops 12\% without CLIP embeddings which could be used for running ROMAN on a more compute-constrained platform.
Removing CLIP embeddings also reduces map size by 100 times.

%% file: paper/limitations.tex
\section{Limitations} 
\label{sec:limitations}

One of the fundamental challenges with using open-set segmentation like FastSAM~\cite{zhao2023fast} for object mapping is determining what constitutes a discrete object. 
ROMAN's filtering and merging steps significantly improve the quality of resulting object maps; however, inconsistent segmentations may sometimes still result in duplicate representations of objects (e.g., a car and each of its doors may be represented as distinct 3D segments).

Additionally, ROMAN seeks to reject non-object-like segments (\eg, ground, walls, etc.) because they do not fit well into the centroid-focused object data association. 
This does not exploit the information present in non-object segments, \eg, roads, walls, and buildings.
Our object registration could additionally be improved by employing a coarse-to-fine technique for using more precise information than object centroids for submap registration.

Finally, while ROMAN runs fast enough for the scale of experiments shown in this paper (\ie, up to eight \SI{1000}{m} long robot trajectories), trajectories longer than this would require significant computation to register the growing number of submaps as robots continue mapping.
A faster place recognition stage could improve scalability.

%% file: paper/conclusion.tex
\section{Conclusion} 
\label{sec:conclusion}
This work presented ROMAN, a method for performing global localization in challenging outdoor environments by robust registration of 3D open-set segment maps.
Associations between maps were informed by geometry of 3D segment locations, object shape and semantic attributes, and the direction of the gravity vector in object maps, 
which enabled global localization even in instances of robots viewing scenes from opposite directions.